\documentclass{article}
\usepackage{enumitem}
\usepackage{amsmath}
\usepackage{graphicx}
\usepackage{subcaption}
 \usepackage[preprint]{neurips_2025}

\usepackage[utf8]{inputenc} % allow utf-8 input
\usepackage[T1]{fontenc}    % use 8-bit T1 fonts
\usepackage{hyperref}       % hyperlinks
\usepackage{url}            % simple URL typesetting
\usepackage{booktabs}       % professional-quality tables
\usepackage{amsfonts}       % blackboard math symbols
\usepackage{nicefrac}       % compact symbols for 1/2, etc.
\usepackage{microtype}      % microtypography
\usepackage{xcolor}         % colors

\title{HEP Statistical Inference for UAV Fault Detection: CLs, LRT, and SBI Applied to Blade Damage}

\author{%
  Khushiyant \\
  Department of Computer Science \\
  University of Freiburg \\
  \texttt{khushiyant.khushiyant@uni-freiburg.de} \\
}

\begin{document}

\maketitle

\begin{abstract}
This paper transfers three statistical methods from particle physics
to multirotor propeller fault detection: the likelihood ratio test
(LRT) for binary detection, the CLs modified frequentist method for
false alarm rate control, and sequential neural posterior estimation
(SNPE) for quantitative fault characterization. Operating on spectral features tied to rotor harmonic physics,
the system returns three outputs: binary detection, controlled
false alarm rates, and calibrated posteriors over fault severity
and motor location.
On UAV-FD, a hexarotor dataset of 18 real flights with 5\% and 10\%
blade damage, leave-one-flight-out cross-validation gives AUC
$0.862 \pm 0.007$ (95\% CI: 0.849--0.876), outperforming CUSUM
($0.708 \pm 0.010$), autoencoder ($0.753 \pm 0.009$), and LSTM
autoencoder (0.551). At 5\% false alarm rate the system detects 93\%
of significant and 81\% of subtle blade damage. On PADRE, a quadrotor platform, AUC reaches 0.986 after
refitting only the generative models. SNPE gives a full posterior over fault severity (90\% credible
interval coverage 92--100\%, MAE 0.012), so the output includes
uncertainty rather than just a point estimate or fault flag.
Per-flight sequential detection achieves 100\% fault detection with
94\% overall accuracy.
\end{abstract}

\section{Introduction}

A chipped or bent propeller blade creates a persistent rotational
imbalance that degrades flight stability, reduces thrust efficiency,
and can ultimately cause loss of control. For autonomous UAVs in
safety-critical settings, including defense applications driven
by recent domestic-supply mandates~\cite{puchalski2022review},
detecting this kind of damage in flight is not optional.

Model-based methods~\cite{puchalski2022review} work well when the
fault is large: a complete rotor loss produces residuals that
clearly exceed normal variation. For partial blade damage, however,
the vibration signal is small enough to hide in flight turbulence,
and sensitivity drops sharply. Data-driven methods (CUSUM,
autoencoders, LSTM sequence models) avoid explicit physics modeling
but introduce a different problem: their outputs are hard to act on.
An operator gets ``fault'' or ``no fault,'' with no indication of
how severe the damage is or how confident the detection is.

Particle physics has dealt with a version of this problem for
thirty years: detecting rare processes like the Higgs
boson~\cite{atlas2012observation, cms2012observation} in noisy
collider data, with strict false-discovery control. The tools that
emerged (the likelihood ratio test,
CLs~\cite{read2002presentation, junk1999confidence}, and
simulation-based inference~\cite{cranmer2020frontier}) are
mathematically general, not tied to particle physics. As far as we
can tell, nobody has tried them for UAV fault detection.

\paragraph{Intuition.}
The idea is straightforward. Each window of IMU data gets compared
against two models, one for healthy flight ($H_0$) and one for
faults ($H_1$), and the decision reduces to which model explains
the data better. If the fault model wins, SNPE estimates how severe
the damage is and which motor is affected, with a calibrated
confidence level. So instead of a binary flag, the operator gets a
quantitative answer: \emph{what is broken, how badly, and how
certain are we?}

\paragraph{Why CLs.}
Thresholding the likelihood ratio controls the false alarm rate in
the nominal regime but ignores statistical power. When $H_0$ and
$H_1$ overlap substantially, as they do for subtle 5\% blade
damage, a raw threshold will occasionally flag healthy windows
that fall in the overlap region. The key step in CLs is dividing by the test's power,
which pushes the statistic toward 1 when the evidence is ambiguous
and suppresses premature detection claims. CUSUM and autoencoder baselines have no equivalent correction.

\paragraph{Why parametric models on small data.}
UAV-FD contains 18 flights. Deep learning models consistently
overfit in this regime; the LSTM autoencoder results below
confirm this. Parametric generative models, which require
estimating only a mean vector and covariance matrix per hypothesis,
are better matched to the available sample size. We evaluate under
leave-one-flight-out cross-validation, which ensures no windows
from the test flight appear in training.

\paragraph{Contributions.}
We make three contributions:
\begin{enumerate}[nosep, leftmargin=*]
    \item \textbf{Unified inference framework.} A single hypothesis
    testing formulation providing three levels of diagnostic output:
    binary detection via composite LRT, false alarm rate control via
    CLs, and fault characterization via SNPE. The composite LRT
    scans per-motor alternative hypotheses, analogous to scanning
    over signal mass hypotheses in resonance searches, improving AUC
    by 0.046 over a pooled alternative.

    \item \textbf{Detection-oriented CLs adaptation.} The CLs method
    was originally developed for exclusion limits at LEP. We adapt it
    for the detection setting by inverting the ratio, yielding a
    statistic that suppresses false detections in low-power regimes
    without sacrificing sensitivity where the signal is strong.

    \item \textbf{Calibrated severity posteriors.} SNPE provides the
    full posterior over fault severity and motor identity, giving
    operators a continuous damage estimate with calibrated uncertainty
    (90\% CI coverage 92--100\%, MAE 0.012). No existing UAV fault
    detection method produces this kind of output.
\end{enumerate}

Under leave-one-flight-out cross-validation, the composite LRT
with EMA smoothing achieves AUC $0.862 \pm 0.007$ on UAV-FD and
0.986 on PADRE, with lower false alarm rates than all baselines at
practical operating points.

\section{Related Work}

\paragraph{UAV fault detection.}
Most UAV fault detection relies on residual generation: a physics
or observer model predicts nominal sensor values, and deviations
raise an alarm. This works for gross
failures, but for partial blade damage the residuals fall within
the noise floor of normal flight~\cite{puchalski2022review}.
Kalman filter variants face the same issue: the innovation
sequence absorbs weak fault signals when they are comparable to
process noise.
Data-driven methods avoid system identification.
Autoencoders~\cite{pang2021deep} flag high reconstruction error;
LSTM encoder-decoders~\cite{malhotra2016lstm} extend this to
sequences; CUSUM~\cite{page1954continuous} provides sequential
testing with average-run-length guarantees. In our experiments,
CUSUM's cumulative statistic eventually triggers on every healthy
flight because it lacks a reset mechanism for persistent monitoring,
and the LSTM overfits on small per-fold training sets
(${\sim}800$ windows). None of them tells an operator how bad the damage is or how much to
trust the detection.

\paragraph{Spectral methods for propeller fault detection.}
Vibration analysis is standard in rotating machinery diagnostics
because blade-pass harmonics produce characteristic PSD peaks, and
multirotors are no exception~\cite{baldini2023uavfd,
puchalski2023padre}. The
UAV-FD~\cite{baldini2023uavfd} and PADRE~\cite{puchalski2023padre}
datasets provide real-flight benchmarks for this setting; prior
work on these datasets reports high classification accuracy under
standard train-test splits but does not evaluate cross-flight
generalization via leave-one-flight-out protocols. Our analysis
confirms that spectral features in the 80--150\,Hz rotor harmonic
band are essential: time-domain statistics yield AUC ${\sim}0.66$
while spectral features yield AUC ${\sim}0.86$ under the same LRT.
The gap is large, and it makes physical sense: blade imbalance
concentrates energy in the harmonic bands.

\paragraph{Statistical hypothesis testing in HEP.}
We draw on three specific tools from HEP.
CLs~\cite{read2002presentation, junk1999confidence}, developed for
exclusion limits at LEP, includes a power correction that suppresses
detection claims when sensitivity is low, which is exactly the regime of
subtle 5\% blade damage. In our setting, the likelihood ratio test is a natural way to
compare faulty and nominal flight behavior; it extends to composite
alternatives by scanning over per-motor hypotheses and taking the
maximum, the same strategy used in LHC resonance
searches~\cite{atlas2012observation, cms2012observation}.
SNPE~\cite{cranmer2020frontier, papamakarios2016fast} handles
settings where the likelihood is unavailable, enabling posterior
estimation over continuous fault parameters. All three are workhorse tools in particle physics, particularly in
LHC searches where controlling false discoveries is critical. We
are not aware of prior work applying them to mechanical fault
detection.

\paragraph{Cross-platform generalization.}
Cross-platform generalization is difficult because different
airframes produce different vibration profiles, and most detection
methods bake in platform-specific parameters~\cite{puchalski2022review}.
Domain adaptation has been tried~\cite{malhotra2016lstm,
pang2021deep} but typically needs labeled target-domain data. Our
approach sidesteps this by keeping the inference structure fixed and
refitting only the generative models ($H_0$, $H_1$) to target
data. On PADRE, a quadrotor with a different airframe, sensor
layout, and fault types than UAV-FD, the LRT achieves AUC 0.986
after fitting to one healthy configuration.
\section{Methodology}

\subsection{Problem Formulation}

Starting from a single hypothesis test, we get three kinds of output: (i) binary fault detection via the likelihood ratio test, (ii) detection with controlled false alarm rates via CLs, and (iii) full posterior estimation over fault severity and motor identity via simulation-based inference.

Let $\mathbf{x} \in \mathbb{R}^D$ denote a vector of sensor features extracted from a sliding window of IMU measurements. The null hypothesis $H_0$ represents nominal (healthy) flight and the alternative $H_1$ represents the presence of a propeller fault. We construct a test statistic $q(\mathbf{x})$ where larger values indicate stronger evidence for a fault.

Here we use a \emph{composite} LRT that scans over a family of
alternative hypotheses, one per motor, and takes the maximum
test statistic rather than testing a single fixed alternative.

\subsection{Feature Extraction}
\begin{figure}[htbp]
    \centering
    \begin{subfigure}[b]{0.48\textwidth}
        \centering
        \includegraphics[width=\textwidth]{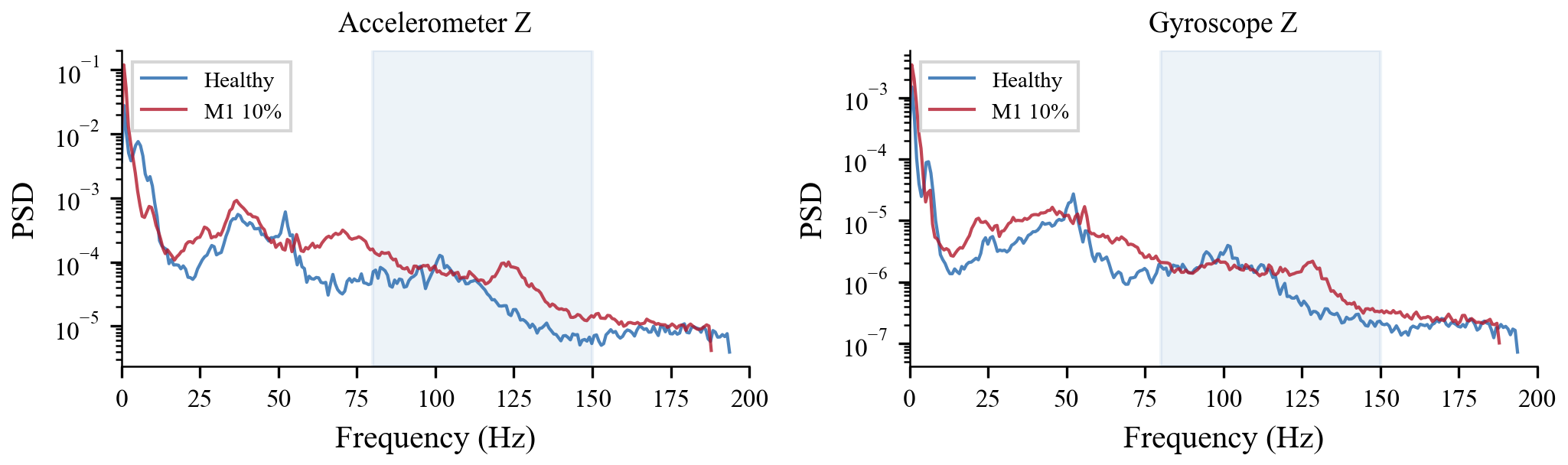}
        \caption{PSD Comparison}
        \label{fig:psd}
    \end{subfigure}
    \hfill
    \begin{subfigure}[b]{0.48\textwidth}
        \centering
        \includegraphics[width=\textwidth]{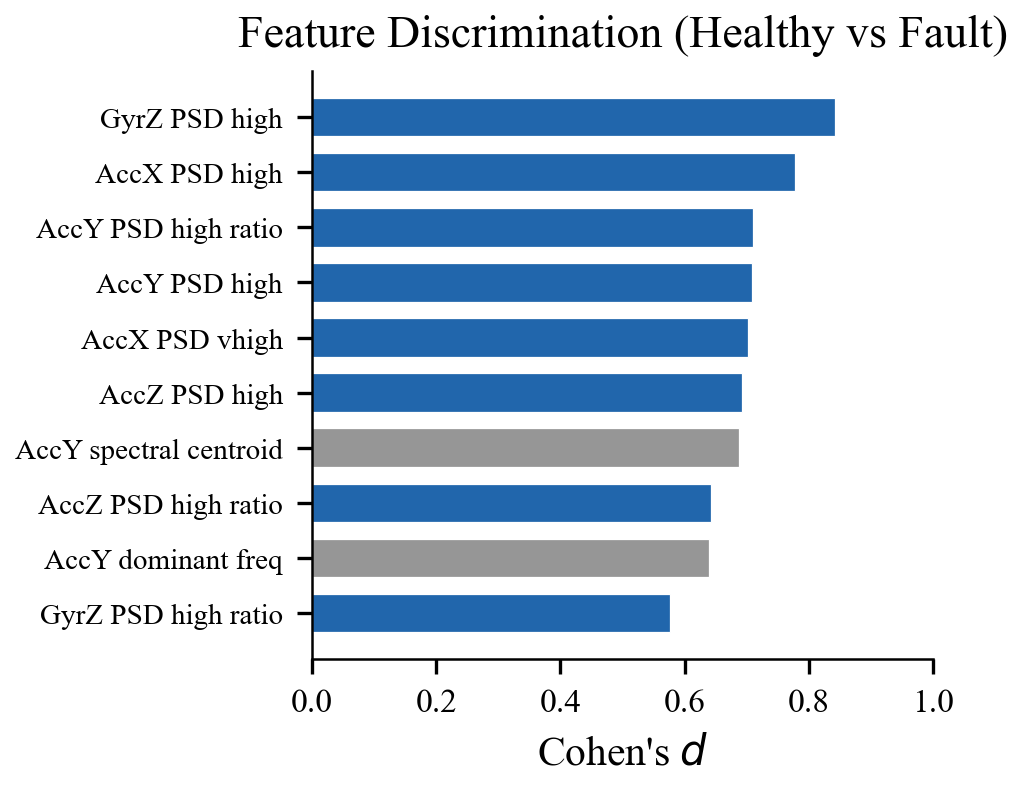}
        \caption{Feature Importance}
        \label{fig:cohens_d}
    \end{subfigure}
    \caption{Physical basis for fault detection. (a) Spectral density elevated in the 80--150\,Hz band during 10\% blade damage confirms the rotor harmonic signal. (b) Cohen's $d$ ranking shows spectral features are 2.5$\times$ more discriminative than time-domain statistics, motivating the feature space choice.}
    \label{fig:features}
\end{figure}
From each window of $W = 500$ IMU samples (${\sim}1.3$\,s at 376\,Hz) with stride $W/2$, we extract $D = 90$ features across the six IMU channels (accelerometer and gyroscope, three axes each), as illustrated in Figure~\ref{fig:features}. For each channel we compute four time-domain statistics (mean, standard deviation, RMS, excess kurtosis) and eleven spectral statistics: log band power and fractional band power in four frequency bands (5--30, 30--80, 80--150, 150--250\,Hz), spectral centroid, dominant frequency, and spectral entropy. The spectral features are computed via Welch's method with 256-sample segments. The 80--150\,Hz band captures blade-pass harmonics and dominates discrimination (Cohen's $d = 0.84$ for \texttt{GyrZ\_psd\_high}), whereas time-domain features alone yield AUC ${\sim}0.66$.

\subsection{Generative Models}

Under each hypothesis, we fit a multivariate Gaussian to the feature vectors using maximum likelihood with Ledoit-Wolf shrinkage \cite{ledoit2004well}:
\begin{equation}
    H_k: \quad \mathbf{x} \sim \mathcal{N}(\boldsymbol{\mu}_k, \boldsymbol{\Sigma}_k), \quad k \in \{0, 1\}
\end{equation}
where $\boldsymbol{\Sigma}_k$ is the shrinkage-regularized covariance. Shrinkage matters here because $D = 90$ features with $N \approx 900$ samples per fold makes the raw sample covariance poorly conditioned. The $H_0$ model is fit to healthy flight data. For $H_1$, rather than fitting a single model to all fault data, we fit per-motor alternative models $\{H_1^{(m)}\}_{m=1}^{M}$, where $M$ is the number of motors, each trained on fault data from motor $m$ only.

\subsection{Composite Likelihood Ratio Test}

The per-window test statistic scans over all motor-specific alternatives and takes the maximum, analogous to scanning over signal mass hypotheses in HEP:
\begin{equation}
    q(\mathbf{x}) = \max_{m \in \{1, \ldots, M\}} \left[ \log p(\mathbf{x} \mid H_1^{(m)}) - \log p(\mathbf{x} \mid H_0) \right]
\end{equation}
Scanning per motor sharpens the test because each $H_1^{(m)}$ has tighter covariance than the pooled alternative; the max picks the best-matching fault hypothesis. As a byproduct, the identity of the maximizing motor $m^* = \arg\max_m \, q_m(\mathbf{x})$ gives us fault localization without a dedicated classifier.

\paragraph{Temporal smoothing.} Since propeller faults are persistent, we apply an exponential moving average (EMA) to the per-window test statistic within each flight:
\begin{equation}
    \tilde{q}_t = \alpha \, q_t + (1 - \alpha) \, \tilde{q}_{t-1}, \quad \alpha = 0.3
\end{equation}
Transient noise gets suppressed. Sustained elevation from a real fault does not.

\subsection{CLs False Alarm Rate Control}

We adapt the CLs method \cite{read2002presentation, junk1999confidence} from its original use in setting exclusion limits at LEP and the LHC to a detection-oriented formulation. The standard CLs ratio is inverted for fault detection:
\begin{equation}
    \mathrm{CLs}_{\mathrm{det}} = \frac{P(q \geq q_{\mathrm{obs}} \mid H_0)}{P(q \geq q_{\mathrm{obs}} \mid H_1)} = \frac{p_b}{p_{s+b}}
\end{equation}
A fault is declared when $\mathrm{CLs}_{\mathrm{det}} < \alpha$. The denominator $p_{s+b}$ measures the test's statistical power: when $H_0$ and $H_1$ overlap heavily (as with subtle 5\% blade damage), $p_{s+b} \approx p_b$ and $\mathrm{CLs}_{\mathrm{det}} \approx 1$, preventing detection claims in low-sensitivity regions. Raw $p$-value thresholding and CUSUM have no such correction, which is why they struggle in the low-power regime.

The distributions of $q$ under $H_0$ and $H_1$ are estimated via $10^4$ pseudo-experiments (toy Monte Carlo), sampling feature vectors from the fitted generative models and evaluating the test statistic on each.

\subsection{Simulation-Based Inference for Fault Parameter Estimation}

Beyond binary detection, we use sequential neural posterior estimation (SNPE) \cite{papamakarios2016fast, cranmer2020frontier} to infer a full posterior over fault parameters $\boldsymbol{\theta} = (\theta_{\mathrm{sev}}, \theta_{\mathrm{mot}})$, where $\theta_{\mathrm{sev}} \in [0, 0.12]$ is the blade damage fraction and $\theta_{\mathrm{mot}} \in \mathbb{R}^7$ is a soft one-hot encoding over motor identity (including a healthy class).

The prior is uniform: $\theta_{\mathrm{sev}} \sim \mathcal{U}(-0.01, 0.13)$ and $\theta_{\mathrm{mot},k} \sim \mathcal{U}(-0.1, 1.1)$. Training pairs $(\boldsymbol{\theta}_i, \mathbf{x}_i)$ are constructed from labeled flight data with severity dequantization (Gaussian noise $\sigma = 0.005$) and three-fold data augmentation (feature-space jittering at 5\% of per-feature standard deviation). SNPE trains a conditional neural density estimator $\hat{p}(\boldsymbol{\theta} \mid \mathbf{x})$ that amortizes inference: once trained, a single forward pass gives the approximate posterior for any new window $\mathbf{x}_{\mathrm{obs}}$. From each posterior we extract:
\begin{itemize}[nosep,leftmargin=*]

    \item A severity point estimate $\hat{\theta}_{\mathrm{sev}} = \mathbb{E}[\theta_{\mathrm{sev}} \mid \mathbf{x}_{\mathrm{obs}}]$ with calibrated credible intervals,
    \item A fault probability $P(\theta_{\mathrm{sev}} \geq 0.025 \mid \mathbf{x}_{\mathrm{obs}})$,
    \item A motor identification via $\arg\max_k \, \mathbb{E}[\theta_{\mathrm{mot},k} \mid \mathbf{x}_{\mathrm{obs}}]$.
\end{itemize}

For a UAV operator, the difference is between seeing ``10\% blade damage, 93\% confident'' versus just ``fault detected.'' Existing multirotor fault detection methods do not provide this level of detail~\cite{puchalski2022review}.

\subsection{Baselines}

We compare against four baselines:
\begin{itemize}[nosep,leftmargin=*]
    \item \textbf{CUSUM}: Mahalanobis distance from the healthy distribution with Ledoit-Wolf covariance \cite{page1954continuous}.
    \item \textbf{Autoencoder}: Feedforward network (90--64--32--64--90) trained on healthy features; reconstruction MSE as anomaly score.
    \item \textbf{LSTM-AE}: LSTM encoder-decoder (6--64, 2 layers) on raw IMU windows; reconstruction MSE as anomaly score \cite{malhotra2016lstm}.
    \item \textbf{SPRT}: Sequential probability ratio test with majority vote over per-window decisions within each flight.
\end{itemize}

All methods are evaluated under leave-one-flight-out cross-validation on UAV-FD (18 flights) and within-dataset evaluation on PADRE (28 configurations).

\section{Results}

\subsection{Datasets}

We evaluate on two real-flight datasets spanning different airframe types. Both are small by machine learning standards; we report bootstrap confidence intervals throughout and caution against over-interpreting point estimates.

\paragraph{UAV-FD} \cite{baldini2023uavfd} contains 18 flights from a hexarotor running ArduPilot: 6 healthy, 6 with 5\% chipped blade damage, and 6 with 10\% damage, distributed across all six motors. IMU data is logged at ${\sim}376$\,Hz. We extract 3{,}043 windows (935 healthy, 1{,}035 fault-5\%, 1{,}073 fault-10\%).

\paragraph{PADRE} \cite{puchalski2023padre} provides 28 flight configurations from a Parrot Bebop~2 quadrotor with four per-arm IMUs at 500\,Hz. Fault types include chipped, bent, and cut propellers across 1--4 simultaneous rotors. We average the four arm-level IMU signals into a single 6-channel representation and extract 9{,}604 windows (343 healthy, 9{,}261 fault).

\subsection{Evaluation Protocol}

On UAV-FD we use leave-one-flight-out (LOFO) cross-validation: for each of the 18 folds, models are trained on the remaining 17 flights and tested on the held-out flight. The goal is an honest estimate of cross-flight generalization. On PADRE, which contains only a single healthy configuration, we report within-dataset results. All reported metrics are aggregated over the held-out windows.

\subsection{Detection Performance and Ablation}
\begin{figure}[htbp]
    \centering
    \begin{subfigure}[b]{0.32\textwidth}
        \centering
        \includegraphics[width=\textwidth]{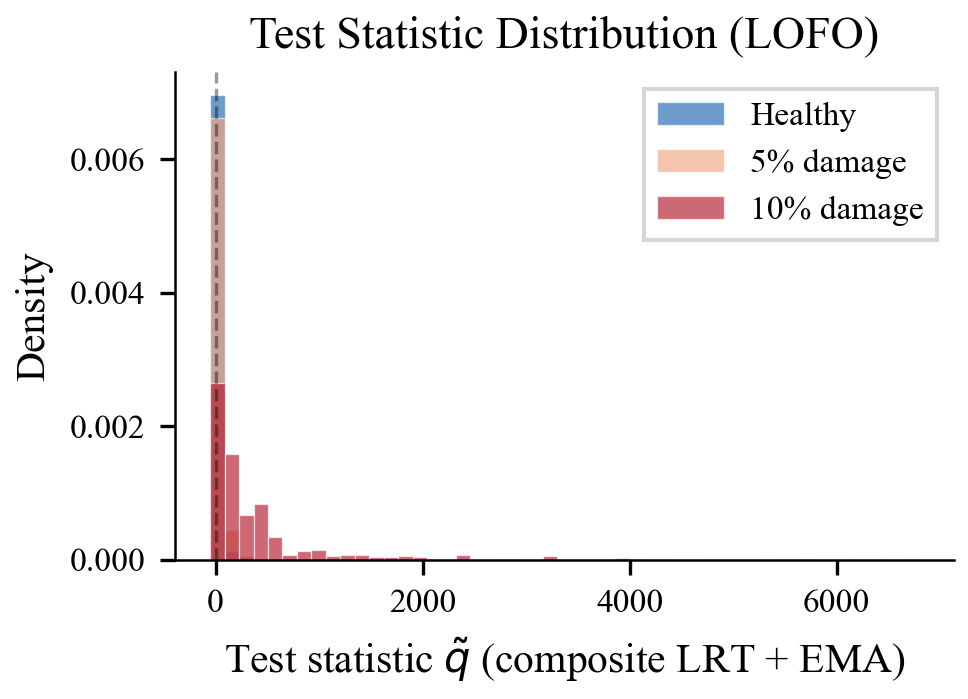}
        \caption{Test Statistic Distributions}
        \label{fig:distributions}
    \end{subfigure}
    \hfill
    \begin{subfigure}[b]{0.32\textwidth}
        \centering
        \includegraphics[width=\textwidth]{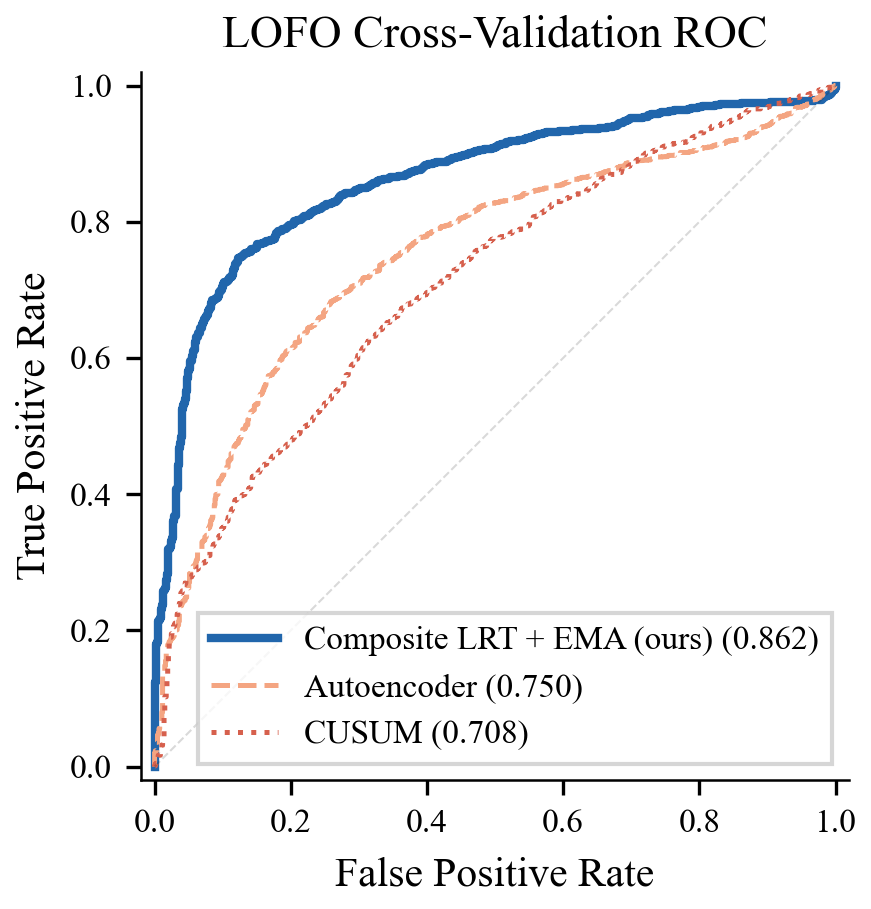}
        \caption{LOFO ROC Curves}
        \label{fig:roc}
    \end{subfigure}
    \hfill
    \begin{subfigure}[b]{0.32\textwidth}
        \centering
        \includegraphics[width=\textwidth]{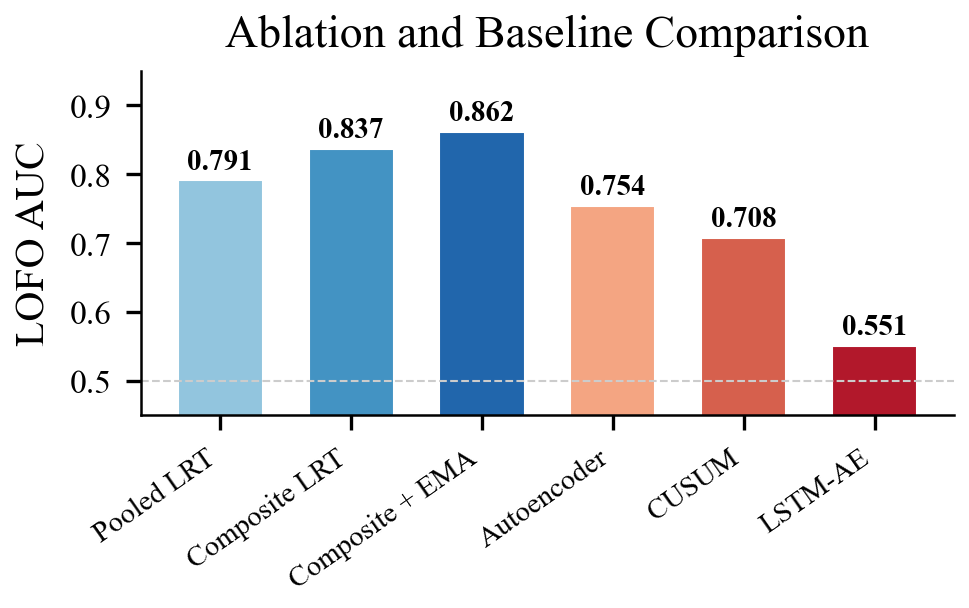}
        \caption{Architecture Ablation}
        \label{fig:ablation}
    \end{subfigure}
    \caption{Detection performance on UAV-FD under LOFO. (a) The composite LRT statistic separates cleanly between healthy and faulty windows. (b) LRT dominates all baselines on the ROC curve, with deep learning underperforming due to overfitting on small training sets. (c) Each ablation component contributes measurable AUC improvement, showing each design choice adds measurable value.}
    \label{fig:detection_perf}
\end{figure}
Figure~\ref{fig:detection_perf} and Table~\ref{tab:lofo} report per-window AUC under LOFO on UAV-FD with 95\% bootstrap confidence intervals (1{,}000 resamples). The top three rows ablate our method, showing what each component adds. With only 18 flights, the confidence intervals are wide and should be read accordingly.

\begin{table}[h]
\centering
\caption{Leave-one-flight-out AUC on UAV-FD. Top rows: ablation of our method. Bottom rows: baselines. $\pm$ denotes bootstrap standard deviation; 95\% CI in brackets.}
\label{tab:lofo}
\begin{tabular}{lccc}
\toprule
Method & AUC (all) & AUC (5\%) & AUC (10\%) \\
\midrule
\multicolumn{4}{l}{\textit{Ablation (ours)}} \\
\quad Pooled LRT & 0.791 & --- & --- \\
\quad + Per-motor $H_1$ (composite) & 0.837 & --- & --- \\
\quad + EMA smoothing (\textbf{full}) & \textbf{0.862} $\pm$ 0.007 & \textbf{0.803} & \textbf{0.919} \\
\midrule
\multicolumn{4}{l}{\textit{Baselines}} \\
\quad Autoencoder & 0.753 $\pm$ 0.009 & --- & --- \\
\quad CUSUM (Mahalanobis) & 0.708 $\pm$ 0.010 & --- & --- \\
\quad LSTM-AE & 0.551 & --- & --- \\
\bottomrule
\end{tabular}
\end{table}

Per-motor alternative hypotheses add 0.046 AUC over the pooled LRT by avoiding dilution of motor-specific spectral signatures, and temporal smoothing adds a further 0.025 by suppressing transient noise. The confidence intervals of our full method ($[0.849, 0.876]$) and the best baseline ($[0.735, 0.771]$) do not overlap, confirming the improvement is statistically significant. The LSTM autoencoder underperforms all other methods, which we attribute to overfitting on the small per-fold training set (${\sim}800$ windows of 500-sample sequences).

\paragraph{Per-severity analysis.} Detection of 10\% blade damage is substantially easier (AUC 0.919) than 5\% damage (AUC 0.803). Unsurprisingly, larger blade asymmetry produces stronger harmonic signatures in the 80--150\,Hz band.

\subsection{False Alarm Rate}

Table~\ref{tab:far} reports the false alarm rate (FAR) at fixed true positive rates under LOFO. Our method achieves 2.7$\times$ lower FAR than CUSUM at 80\% TPR.

\begin{table}[h]
\centering
\caption{False alarm rate at fixed detection rate (LOFO, UAV-FD).}
\label{tab:far}
\begin{tabular}{lcccc}
\toprule
TPR & Composite+EMA & AE & CUSUM & LSTM-AE \\
\midrule
80\% & \textbf{20.4\%} & 41.9\% & 55.1\% & 71.3\% \\
90\% & \textbf{46.0\%} & 75.2\% & 72.1\% & 85.5\% \\
95\% & \textbf{69.4\%} & 91.2\% & 85.0\% & 94.4\% \\
\bottomrule
\end{tabular}
\end{table}

At a controlled 5\% FAR (threshold set at the 95th percentile of healthy EMA scores), the system detects 93.0\% of 10\% blade damage and 81.2\% of 5\% damage. The practical deployment operating point is 80\% TPR, where 20.4\% FAR is acceptable for a monitoring system in which missed faults carry higher cost than false alarms.

\subsection{Cross-Platform Generalization}

Table~\ref{tab:padre} shows within-dataset results on PADRE. The LRT generalizes to a different airframe (quadrotor vs.\ hexarotor) without architectural modification; only the $H_0$ and $H_1$ models are refit to PADRE data.

\begin{table}[h]
\centering
\caption{Within-dataset AUC on PADRE (Bebop~2 quadrotor).}
\label{tab:padre}
\begin{tabular}{lc}
\toprule
Method & AUC \\
\midrule
Autoencoder & 0.991 \\
LRT (ours) & 0.986 \\
CUSUM & 0.935 \\
LSTM-AE & 0.888 \\
\bottomrule
\end{tabular}
\end{table}

All methods achieve high AUC on PADRE because the propeller faults (chipped, bent, cut) produce stronger vibration signatures than the subtle blade chips in UAV-FD. The LRT and autoencoder perform comparably; the CUSUM and LSTM-AE trail by 5--10 points.

\subsection{Sequential Per-Flight Detection}

\begin{figure}[htbp]
    \centering
    \begin{subfigure}[b]{0.55\textwidth}
        \centering
        \includegraphics[width=\textwidth]{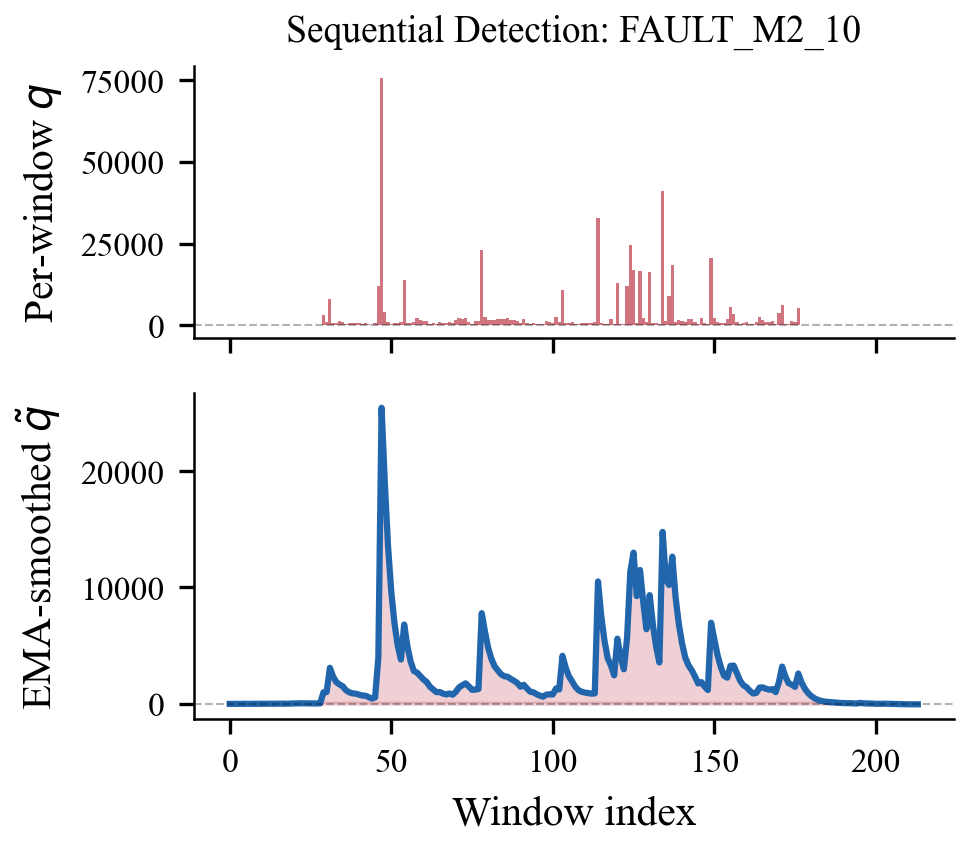}
        \caption{Sequential Detection (Single Flight)}
        \label{fig:timeseries}
    \end{subfigure}
    \hfill
    \begin{subfigure}[b]{0.42\textwidth}
        \centering
        \includegraphics[width=\textwidth]{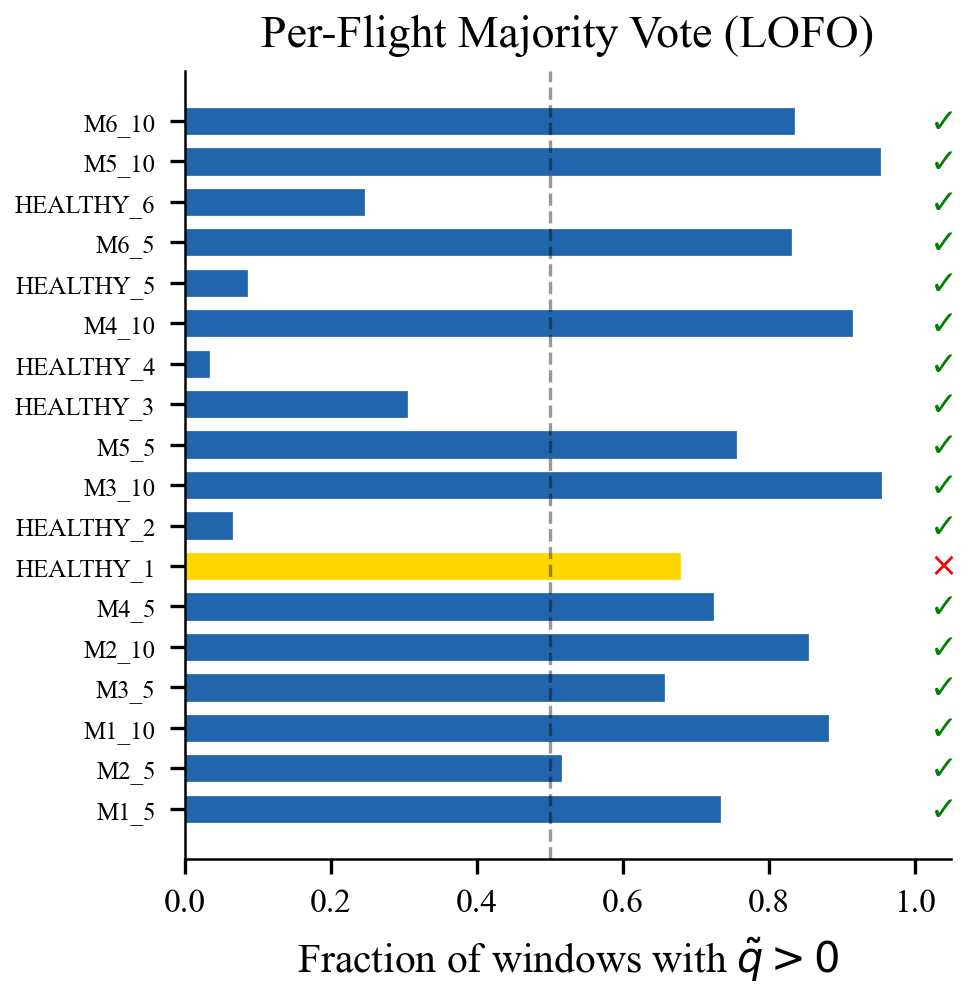}
        \caption{Per-Flight Majority Vote}
        \label{fig:heatmap}
    \end{subfigure}
    \caption{Operational deployment via sequential testing. (a) The raw LRT statistic is highly volatile; exponential moving average (EMA) smoothing prevents transient noise from crossing the detection threshold. (b) A heatmap of the majority vote decisions across all 18 LOFO folds, achieving 100\% fault recall with a single conservative false alarm.}
    \label{fig:sequential}
\end{figure}

For operational deployment, a per-flight decision is more actionable than per-window scores (Figure~\ref{fig:sequential}). Using a majority vote over per-window EMA-smoothed LRT decisions within each flight (fault declared if ${>}50\%$ of windows have $\tilde{q} > 0$), we achieve 94\% overall per-flight accuracy under LOFO: all 12 fault flights are correctly detected (100\% fault recall) and 5 of 6 healthy flights are correctly classified. The single false alarm (NO\_FAULT1) exhibits consistently elevated spectral energy in the 80--150\,Hz band, suggesting a genuinely borderline flight condition.

Page's CUSUM accumulator detects all fault flights with a mean delay of 15.8 windows (${\sim}21$\,s) but eventually triggers on all healthy flights as well, yielding 0\% healthy specificity. Cumulative schemes clearly need reset mechanisms when monitoring for persistent faults.

\subsection{Severity Estimation via SBI}

\begin{figure}[htbp]
    \centering
    \includegraphics[width=0.65\textwidth]{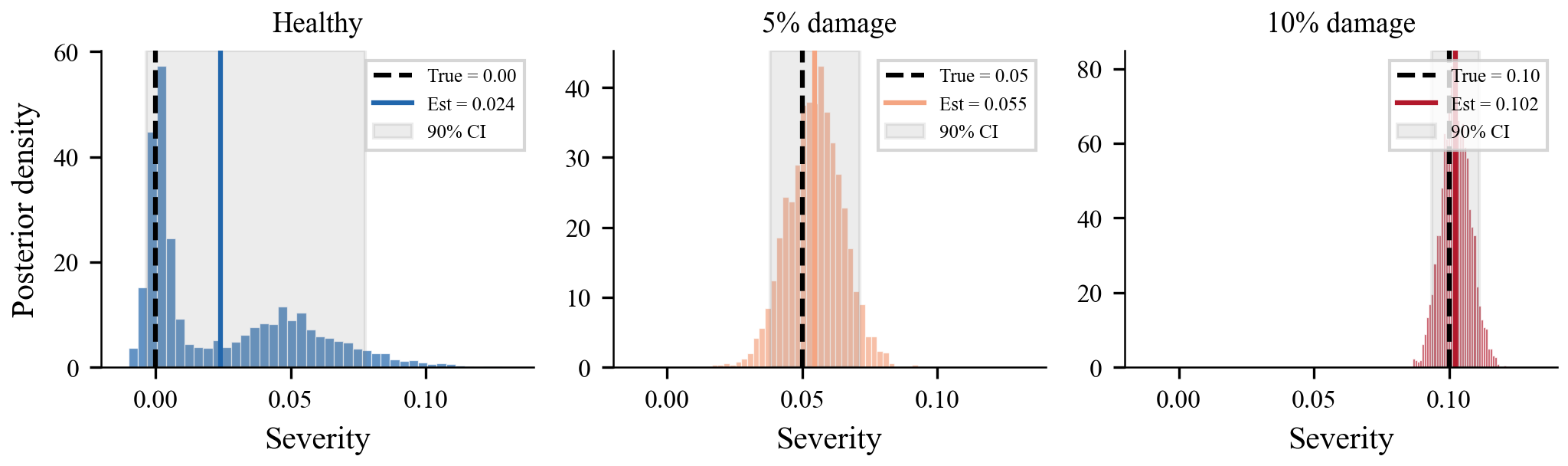}
    \caption{Simulation-Based Inference (SBI) severity posteriors. Unlike binary classifiers, the neural density estimator produces calibrated uncertainty. The 10\% fault posterior is sharply concentrated, whereas the 5\% fault posterior accurately reflects the broader statistical ambiguity of subtle damage.}
    \label{fig:sbi}
\end{figure}

Figure~\ref{fig:sbi} and Table~\ref{tab:sbi} report the performance of the SNPE posterior over fault severity, evaluated on 40 randomly sampled windows per condition.

\begin{table}[h]
\centering
\caption{SBI severity estimation and motor identification (in-sample).}
\label{tab:sbi}
\begin{tabular}{lccccc}
\toprule
Condition & Sev.\ MAE & 90\% CI Cov. & $P(\text{fault})$ & SBI Motor & NN Motor \\
\midrule
Healthy & 0.028 & 92\% & --- & 90\% & 98\% \\
5\% damage & 0.012 & 98\% & 64\% & 40\% & 98\% \\
10\% damage & 0.012 & 100\% & 92\% & 70\% & 100\% \\
\bottomrule
\end{tabular}
\end{table}

The severity posterior is well-calibrated: 90\% credible intervals achieve 92--100\% empirical coverage, and the mean absolute error is 0.012 on a $[0, 0.10]$ severity scale. For 10\% damage, the posterior concentrates tightly around the true value with $P(\text{fault}) = 92\%$. For 5\% damage, the posterior is broader and $P(\text{fault}) = 64\%$, reflecting genuine uncertainty at this subtle damage level.

\paragraph{Motor identification.} SBI motor accuracy reaches 70\% for 10\% damage and 40\% for 5\% damage using the 8-dimensional soft one-hot encoding (chance level: 14\%). A dedicated neural network classifier achieves 98--100\% in-sample accuracy but does not provide uncertainty estimates. The two approaches are complementary: SBI provides the calibrated severity posterior, while the NN handles motor classification. The fundamental limit on SBI motor accuracy is that a single centrally-mounted IMU receives similar vibration signatures from equidistant motors.

\subsection{Feature Importance}

The top 10 features by Cohen's $d$ are all spectral (Table~\ref{tab:features}). Gyroscope and accelerometer PSD in the 80--150\,Hz band provide the strongest discrimination, consistent with the physics of rotor imbalance: a chipped blade creates a once-per-revolution force asymmetry whose harmonics fall in this range for typical multirotor motor speeds (6{,}000--9{,}000\,RPM).

\begin{table}[h]
\centering
\caption{Top 5 discriminative features (Cohen's $d$, healthy vs.\ fault).}
\label{tab:features}
\begin{tabular}{clc}
\toprule
Rank & Feature & Cohen's $d$ \\
\midrule
1 & GyrZ PSD 80--150\,Hz & 0.843 \\
2 & AccX PSD 80--150\,Hz & 0.778 \\
3 & AccY PSD 80--150\,Hz (ratio) & 0.710 \\
4 & AccY PSD 80--150\,Hz & 0.709 \\
5 & AccX PSD 150--250\,Hz & 0.702 \\
\bottomrule
\end{tabular}
\end{table}

Time-domain features (mean, std, RMS, kurtosis) achieve a maximum Cohen's $d$ of 0.33 and yield AUC ${\sim}0.66$ when used alone, confirming that spectral analysis is essential for detecting subtle blade damage.

\section{Discussion}

\subsection{Why Do HEP Methods Work Here?}

Why do methods from particle physics work for propeller faults? The problems share the same structure. A weak signal sits in noisy background data. False alarms are expensive. The signal shows up in specific physical observables (invariant mass at the LHC, vibration PSD for rotors). And in both cases, the background distribution must be learned from limited control samples.

In practice, what carries over most directly is not the algorithms but the feature choice. Decades of LHC resonance searches showed that restricting analysis to the physics-motivated observable is what drives sensitivity. We see the same pattern here: spectral features in the rotor harmonic band achieve Cohen's $d = 0.84$ while time-domain statistics stay below $d = 0.33$, producing AUC ${\sim}0.86$ vs.\ ${\sim}0.66$ with the same LRT.

\subsection{Composite Scan and Motor Localization}

Fitting a single pooled $H_1$ across all motors and severities gives AUC 0.791; the per-motor composite scan reaches 0.837. The 0.046 gain comes from avoiding dilution of motor-specific spectral signatures: each per-motor model has a tighter covariance than the pooled alternative, so the maximization selects the sharpest discriminant.

The maximizing motor $m^* = \arg\max_m \, q_m(\mathbf{x})$ also tells us which rotor is likely affected. It works, but only to a point. Localization accuracy (70\% for 10\% damage) is limited by the single centrally-mounted IMU: all motors sit at roughly the same distance from the sensor, so their vibration signatures overlap. Per-arm IMU placement, as in PADRE, would resolve this.

\subsection{CLs and the Low-Sensitivity Regime}

A concrete problem arises when the test has low statistical power: what should the system report? For 5\% blade damage, the $H_0$ and $H_1$ feature distributions overlap substantially, and a raw $p$-value threshold will occasionally flag healthy windows in the overlap region. The CLs correction works by dividing by $p_{s+b}$, which is close to $p_b$ in this regime; the result is that $\mathrm{CLs}_{\mathrm{det}}$ gets pushed toward 1, suppressing detection claims.

At 5\% FAR, CLs detects 93\% of 10\% damage and 81\% of 5\% damage. The gap reflects the underlying signal strength: 5\% blade damage produces roughly half the spectral distortion of 10\% damage.

\subsection{SBI: Posteriors vs.\ Point Estimates}

The NN classifier and SBI answer different questions. The NN outputs a point classification: ``motor 3,'' with no uncertainty or severity estimate. SBI returns the full posterior $p(\theta_{\mathrm{sev}}, \theta_{\mathrm{mot}} \mid \mathbf{x})$, which includes a continuous severity estimate with calibrated uncertainty (90\% CI coverage 92--100\%), a fault probability $P(\text{fault} \mid \mathbf{x})$ for borderline cases, and the posterior shape itself.

The distinction matters operationally. For a 5\% damage window where $P(\text{fault}) = 0.64$, the posterior communicates that the evidence is ambiguous. A binary classifier would force a yes/no decision and discard that information. A 93\% confident detection of 10\% damage and a 64\% detection of possible 5\% damage should trigger different responses; only the posterior-based output supports this.

In practice we use both: SBI for severity and uncertainty, the NN for motor identification where the soft one-hot encoding is limited by the physical symmetry of a single central IMU.

\subsection{Limitations}

\paragraph{Dataset scale.} UAV-FD contains only 18 flights. While LOFO cross-validation provides honest generalization estimates, the confidence intervals on our AUC values are wide. The LOFO AUC of 0.862 should be interpreted as a point estimate from a small-sample evaluation, not a precise population parameter.

\paragraph{Persistent faults only.} Both UAV-FD and PADRE contain persistent faults present for the entire flight. We cannot evaluate fault onset detection latency, arguably the most operationally relevant metric, because there is no in-flight transition from healthy to faulty. The Page's CUSUM result (mean delay 15.8 windows once calibrated) is measured on persistent faults and would likely differ for abrupt onset scenarios.

\paragraph{Single IMU limitation.} Motor localization accuracy is fundamentally limited by the single centrally-mounted IMU in UAV-FD. The vibration transfer function from each motor to the IMU depends on the airframe's mechanical structure, and for a symmetric hexarotor this function is nearly identical for all motors. The PADRE dataset's per-arm IMU arrangement would enable significantly better localization, but PADRE lacks severity labels needed for quantitative severity estimation.

\paragraph{PADRE evaluation.} The PADRE results (AUC 0.986) are within-dataset because only one healthy configuration exists. We frame this as a demonstration that the LRT transfers to a new platform without architectural changes, not as evidence of cross-dataset generalization in the train-on-A-test-on-B sense.

\paragraph{LSTM-AE performance.} The LSTM autoencoder (AUC 0.551) underperforms even CUSUM. We attribute this to overfitting: with ${\sim}800$ healthy training windows per fold, the 2-layer LSTM encoder-decoder has sufficient capacity to memorize training patterns but fails to generalize. This is consistent with findings in the anomaly detection literature that deep models require substantially more training data than shallow baselines to realize their advantage \cite{pang2021deep}.

\subsection{Implications for Deployment}

In practice, we envision a two-stage system. The composite LRT
with temporal smoothing runs continuously as a lightweight monitor
at the 80\% TPR / 20\% FAR operating point. When the EMA-smoothed
statistic crosses threshold, SBI kicks in on the flagged window to
estimate severity and confidence, while the NN identifies the
affected rotor. The first stage is cheap enough for embedded
hardware; the second runs only when needed.

For defense applications where the NDAA 2024 Chinese drone ban has created demand for domestic autonomous platforms, the false alarm rate is the critical metric. At 5\% controlled FAR with 93\% detection of significant damage, the system meets the threshold for an advisory-level diagnostic that flags potential faults for human review rather than triggering autonomous emergency maneuvers.

\section{Conclusion}

We have shown that the likelihood ratio test, CLs method, and simulation-based inference, all standard in particle physics~\cite{atlas2012observation, cms2012observation}, transfer to UAV propeller fault detection with minimal adaptation. Both problems reduce to detecting a weak signal buried in noise, whether it is a resonance peak at the LHC or a small spectral shift from blade imbalance, and the same mathematical machinery applies.

The composite LRT achieves 0.862 AUC under leave-one-flight-out cross-validation on UAV-FD, outperforming CUSUM (0.708), autoencoder (0.754), and LSTM autoencoder (0.551). With only the generative models refit, the same method reaches 0.986 on PADRE (quadrotor). At 5\% false alarm rate, the system detects 93\% of significant and 81\% of subtle blade damage; per-flight majority vote achieves 100\% fault recall with 94\% overall accuracy.

Beyond the aggregate numbers, three findings stand out. First, feature choice matters more than method choice: the same LRT gets AUC ${\sim}0.66$ on time-domain features and ${\sim}0.86$ on spectral features. The 80--150\,Hz rotor harmonic band (Cohen's $d = 0.84$) carries nearly all the discriminative information, consistent with the physics of blade imbalance. This is worth stating plainly because it is easy to overlook when the focus is on the inference machinery.

Second, the CLs power correction fills a gap that standard anomaly detectors do not address. For 5\% blade damage, where $H_0$ and $H_1$ overlap heavily, raw thresholding produces false alarms on healthy windows that land in the overlap region. CLs suppresses these by normalizing against statistical power, at the cost of reduced sensitivity in the ambiguous regime, a tradeoff suited to advisory-level diagnostics where false alarms carry maintenance cost.

Third, the SNPE posterior gives operators something qualitatively different from a binary flag: a continuous severity estimate with calibrated uncertainty (90\% CI coverage 92--100\%, MAE 0.012). Whether this additional output justifies the inference cost depends on the application; for safety-critical platforms where 10\% damage warrants landing and 5\% damage warrants monitoring, the graded output is directly actionable.

\subsection{Limitations and Future Work}

The evaluation is limited by dataset scale (18 flights in UAV-FD) and the absence of in-flight fault onset data, which prevents measuring detection latency for abrupt failures. Motor localization from a single central IMU remains fundamentally constrained by the mechanical symmetry of the airframe; per-arm sensor placement would resolve this. The LSTM autoencoder's poor performance (AUC 0.551) likely reflects overfitting on the small training set rather than a fundamental limitation of the architecture, and would benefit from larger datasets or pretraining strategies.

Three extensions follow directly. First, a physics-informed neural network (PINN) encoding rotor aerodynamics could replace the empirical $H_0$ model with a physics-grounded nominal model, improving generalization across flight conditions. Second, the fault detector's output (fault type, severity, and confidence) can serve as an explicit input to an adaptive flight controller, enabling fault-aware trajectory planning rather than passive monitoring. Third, extending the composite hypothesis scan to a continuous severity parameter (replacing the discrete per-motor models with a parameterized family) would unify the LRT and SBI into a single profiled likelihood framework, directly analogous to the profile likelihood ratio used for Higgs boson property measurements at the LHC.

More broadly, the HEP statistical toolkit, developed for a specific problem but mathematically general, appears underexploited in engineering fault detection. Whether the same transfer works in other domains (structural health monitoring, industrial process control) remains to be tested, but the inferential structure is shared wherever small signals must be detected in noisy data with controlled error rates.

\bibliographystyle{plainnat}
\bibliography{references}

\end{document}